\documentclass[10pt, a4paper]{article}

\usepackage{graphicx}
\usepackage{inconsolata}
\usepackage{tabularx, booktabs}
\usepackage{enumitem, listings, xcolor}
\usepackage[most]{tcolorbox}
\usepackage{multirow}
\usepackage{hyperref}
\usepackage{ulem}

\usepackage{xeCJK}
\usepackage[final]{lrec2026} 

\title{Emotion Transcription in Conversation: A Benchmark for Capturing Subtle and Complex Emotional States through Natural Language}

\name{Yoshiki Tanaka$^{1}$, Ryuichi Uehara$^{1}$, Koji Inoue$^{2}$, Michimasa Inaba$^{1}$}

\address{
    $^{1}$The University of Electro-Communications, Tokyo, Japan 
    $^{2}$Kyoto University, Kyoto, Japan\\
    \{y-tanaka,r-uehara,m-inaba\}@uec.ac.jp\\
    inoue.koji.3x@kyoto-u.ac.jp
}

\abstract{
Emotion Recognition in Conversation (ERC) is critical for enabling natural human-machine interactions. However, existing methods predominantly employ categorical or dimensional emotion annotations, which often fail to adequately represent complex, subtle, or culturally specific emotional nuances. To overcome this limitation, we propose a novel task named Emotion Transcription in Conversation (ETC). This task focuses on generating natural language descriptions that accurately reflect speakers' emotional states within conversational contexts. To address the ETC, we constructed a Japanese dataset comprising text-based dialogues annotated with participants' self-reported emotional states, described in natural language. The dataset also includes emotion category labels for each transcription, enabling quantitative analysis and its application to ERC. We benchmarked baseline models, finding that while fine-tuning on our dataset enhances model performance, current models still struggle to infer implicit emotional states. The ETC task will encourage further research into more expressive emotion understanding in dialogue.
The dataset is publicly available at \url{https://github.com/UEC-InabaLab/ETCDataset}.
\\ \newline  }

\begin{document}

\maketitleabstract

\section{Introduction}
Emotion Recognition in Conversation (ERC) is a critical technology for enabling natural and seamless human-machine interactions, and its significance has grown substantially in recent years~\cite{poria2019emotion,pereira2025deep}.
Accurately comprehending user emotions is indispensable for developing more empathetic and human-like conversational systems.
Research in this domain has advanced dramatically over the past decade, with the advent of deep learning and large-scale language models propelling recognition accuracy toward practical application levels~\cite{Fu2025LaERCS,pereira2025deep}.
Numerous benchmark datasets have been established~\citelr{Busso2008iemocap,Li2017dailydialog,Poria2019meld}, and emotion classification algorithms have evolved from traditional machine learning to sophisticated deep learning architectures.
However, a prevalent limitation persists: dialogue data predominantly used in existing research often relies on specific scenarios or includes posed emotional expressions, which do not always reflect spontaneous, everyday conversations~\citelr{Busso2008iemocap,Poria2019meld}.
Consequently, this presents a considerable challenge for conversational systems and robots designed for diverse real-world applications.

\begin{figure}[t!]
    \centering
    \includegraphics[width=\linewidth]{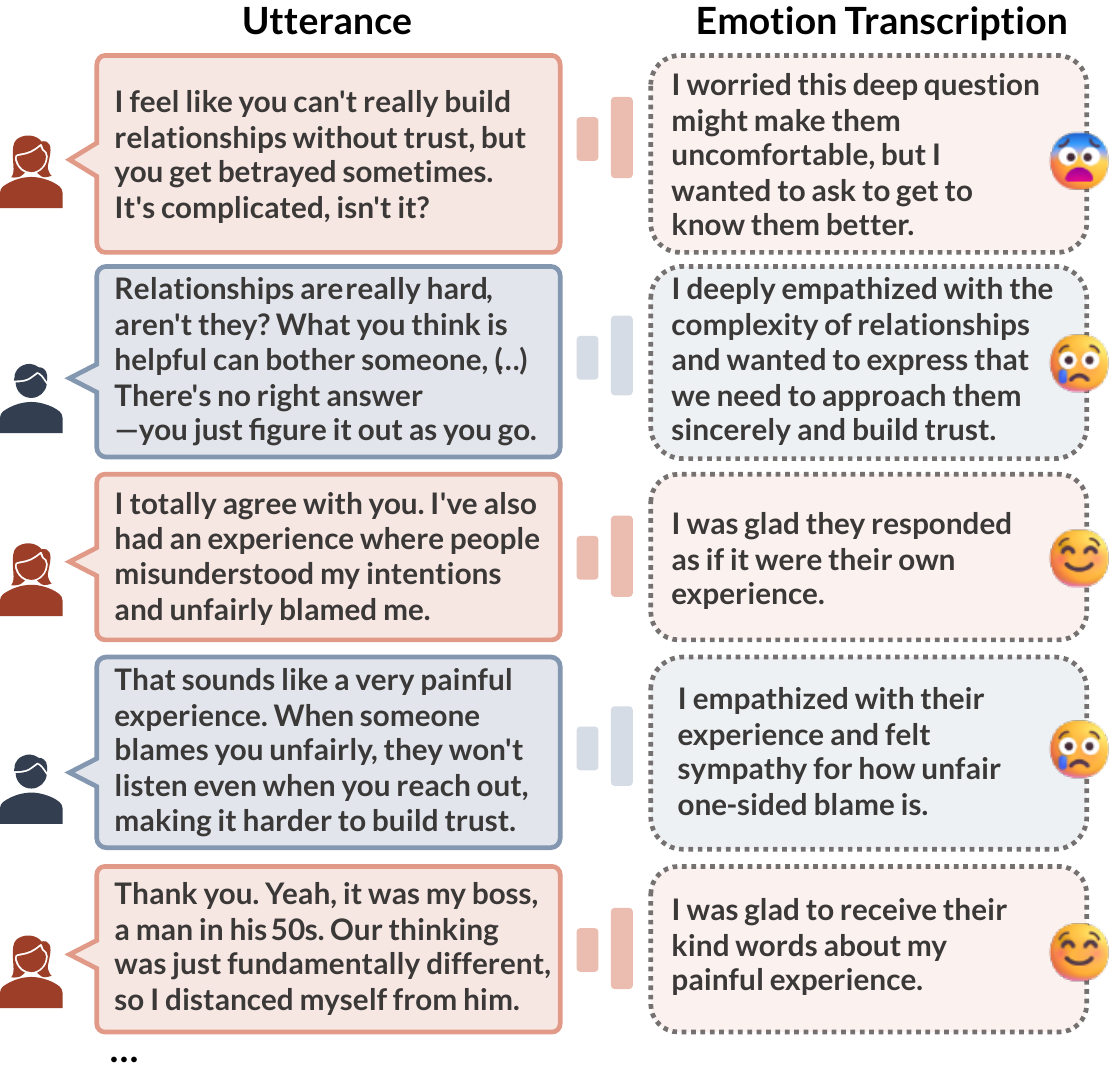}
    \caption{Example of a dialogue with emotion transcriptions from our dataset~(translated from Japanese). Each transcription is also annotated with multi-label emotion categories.}
    \label{fig:example}
\end{figure}

Addressing this gap, we propose a novel task for emotion recognition: ``Emotion Transcription in Conversation (ETC).''
Unlike conventional ERC approaches, ETC aims to generate natural language descriptions of a speaker's emotional state for each utterance within a conversation.
By capturing emotions through natural language, we anticipate the ability to represent richer and more detailed affective information that eludes traditional categorical or numerical frameworks.
This includes subtle emotional shifts, blended emotions, and culturally specific expressions of feeling, which has been attempted by soft label of emotional categories so far~\cite{wen2023learning}.
For instance, complex states such as ``a sense of exasperation mixed with concern for the other person'' or ``disappointment at an unmet expectation while trying to maintain composure'' are difficult to articulate within existing paradigms.
Describing emotions in natural language opens a new avenue for machines to understand and articulate the inherent richness of human affective states.
Although there has been a limited number of attempts to represent emotional states of dialogue participants in natural language~\citelr{shinoda2025tomato}, those were artificially generated by LLMs.
This formulation is particularly exigent in the current era of rapidly advancing LLM capabilities, offering the potential for machines to more accurately grasp the depths of human emotion and thereby foster genuinely empathetic interactions.

To realize the proposed ETC task, our initial undertaking involves the meticulous collection of a suitable dataset and the establishment of a robust benchmark.
Specifically, we leverage crowdsourcing to collect text-based dialogues in Japanese that simulate a variety of everyday scenarios.
Crucially, during this process, dialogue participants will be prompted to describe their own emotional state at each turn of the conversation using free-form natural language, as depicted in Figure~\ref{fig:example}. Furthermore, to enable quantitative analysis and its application to ERC, we collect emotion category labels corresponding to each transcription.
This methodology yields a large-scale dataset comprising pairs of naturalistic emotional expressions and their corresponding linguistic descriptions.
Subsequently, using this curated dataset, we develop and evaluate models designed to generate these emotional state descriptions from dialogue context.
This empirical validation ascertains the feasibility of the ETC task and the efficacy of our proposed approach.

The primary contributions of this research are threefold:
(1) We introduce ``Emotion Transcription in Conversation (ETC)'' as a new task in conversational emotion recognition, enabling a more nuanced and detailed understanding of emotions beyond what is possible with existing categorical or dimensional approaches.
This represents a paradigm shift in how machines can interpret and represent affective states.
(2) We construct and release a novel dialogue dataset annotated with speakers' own natural language descriptions of their emotions.
This dataset will serve as a valuable resource for the research community, particularly for studies aiming to achieve more human-like emotional understanding in AI.
(3) We develop baseline models for the ETC task using this dataset and empirically demonstrate their effectiveness.
In doing so, this research charts a new direction for emotion recognition technology and contributes to the advancement of emotional intelligence in future conversational systems and robotics, paving the way for more sophisticated and empathetic human-AI collaboration.

\section{Related Work}

ERC has garnered significant attention within Natural Language Processing (NLP), driven by the exponential increase in recognition models with benchmark datasets.

\paragraph{Recognition Models}

Pre-trained language models (PLMs), such as BERT~\cite{Devlin2019BERT} and RoBERTa~\cite{Liu2019RoBERTa}, have become foundational in text-based ERC due to their powerful contextual understanding.
Many approaches fine-tune these PLMs on specific ERC datasets, often augmenting them with additional layers to integrate conversational features like speaker identities and dialogue history~\cite{Qin2023BERTERC,Yu2024EmotionAnchored}.
Graph Neural Networks (GNNs) have also gained prominence in modeling intricate conversational dynamics, particularly speaker interactions and utterance dependencies. Notable examples include DialogueGCN~\cite{Ghosal2019DialogueGCN}, which represents conversations as graphs with utterances as nodes and interrelations as edges.
Additionally, models like ESIHGNN have introduced event-state interactions within heterogeneous GNN frameworks to enhance ERC performance~\cite{Zhang2024ESIHGNN}.
Large Language Models (LLMs) have further advanced ERC through diverse prompting strategies, such as zero-shot, few-shot, and chain-of-thought prompting, as well as targeted LoRA fine-tuning~\cite{Feng2024AffectLLM}.
Recent studies like LaERC-S specifically explore enhancing LLM-based ERC through explicit integration of speaker characteristics~\cite{Fu2025LaERCS}.

\paragraph{Datasets and Emotion Categories}

Several benchmark datasets have been developed to advance ERC research, often employing discrete emotion categories.
MELD (Multimodal EmotionLines Dataset), derived from the ``Friends'' TV series, captures multi-party interactions annotated with Ekman's six basic emotions and neutral, making it widely used for studying emotional shifts in conversational contexts~\citelr{Poria2019meld}.
The IEMOCAP (Interactive Emotional Dyadic Motion Capture Database) dataset provides dyadic interactions with both scripted and improvised dialogues, supplemented by multimodal data, and features custom emotional labels such as anger, happiness/excitement, neutral, and sadness~\citelr{Busso2008iemocap}.
MSP-Conversation~\citelr{martinez-etal-2020-msp-conversation} and MSP-Podcast~\citelr{busso-etal-2025-msp-podcast} are audio-based corpora sourced from podcast recordings. MSP-Conversation provides time-continuous annotations of emotional dimensions (arousal, valence, and dominance) over entire conversations, while MSP-Podcast offers segment-level annotations with categorical emotion labels. MSP-Podcast also utilizes typed descriptions as supplementary annotations from annotators when they select the ``other'' emotion category, allowing annotators to specify emotions beyond predefined categories~\cite{chou-etal-2022-exploiting}.
DailyDialog offers cleaner, human-generated dyadic dialogues annotated with Ekman's basic emotions and includes communicative intent labels~\citelr{Li2017dailydialog}.
EmotionLines, also based on ``Friends,'' is a text-only predecessor of MELD~\citelr{Hsu2018EmotionLines}. 
Similarly, EmoryNLP, although also drawn from ``Friends,'' provides distinct data splits and annotations~\citelr{Zahiri2018EmoryNLPDataset}.
EmoContext focuses on short, three-turn dialogues with simplified emotion categories (happiness, sadness, anger, others)~\citelr{chatterjee2019semeval}.
CAPE, a Chinese appraisal-based emotional dataset, specifically targets emotional generation in LLMs~\citelr{liu-etal-2025-cape}.
While some attempts have been made to represent dialogue participants' emotional states explicitly in natural language, similar to our ETC task, these efforts primarily relied on artificially generated data from LLMs~\citelr{shinoda2025tomato}.

\section{Data Collection}

\subsection{Procedure}

We used CrowdWorks\footnote{https://crowdworks.jp}, a Japanese crowdsourcing platform, for this data collection.
To characterize our participant pool, we measured the Big Five personality traits of all crowdworkers using a 10-item questionnaire based on the Japanese version of the Ten-Item Personality Inventory (TIPI-J) \cite{oshio2012development}.
To effectively elicit rich and naturally occurring emotional expressions, we adopted a speaker-listener dialogue setting inspired by the EmpatheticDialogue paradigm~\citelr{rashkin-etal-2019-towards}.
Crowdworkers were assigned one of two roles: a ``Speaker'' or a ``Listener.''
The core task for the Speaker was to recount a personal episode related to a specific emotion. 
To guide this, Speakers were provided with one of 32 emotion labels, which were Japanese translations of the emotions used in the EmpatheticDialogue corpus~\citelr{rashkin-etal-2019-towards}. 
They were instructed to convey, through dialogue, a concrete personal experience where they felt the designated emotion.
Listeners, in turn, were tasked with actively engaging with the Speaker's narrative. 
Dialogues were structured to begin with an utterance from the Speaker, followed by alternating turns between the Speaker and Listener. 
Each dialogue concluded after 5 turns, resulting in a total of 10 utterances.
Immediately after inputting each of their utterances, both the Speaker and the Listener were required to provide a free-form textual description of their own internal emotional state at that specific moment of uttering the text. 
We defined this emotional state description as ``a verbalization of the internal emotional state or intention a participant held at the time of their utterance.''
Crowdworkers were explicitly guided to describe, in as much detail as possible, the psychological context behind their utterance. 
Further details on the data collection are provided in Appendix \ref{sec:appx_collection}.

\subsection{Data Collection Results}
We collected a total of 1,002 dialogues. An example of collected data is presented in Figure~\ref{fig:example}.
A total of 199 unique crowdworkers participated in the data collection. The distribution of personality traits among participants is provided in Appendix~\ref{sec:appx_personality} for reference.
The median number of dialogues per worker was 6.0, with the most active worker contributing to 38 dialogues. 
We aimed for an even distribution of dialogues across the 32 emotion labels; the number of dialogues per emotion ranged from a minimum of 30 to a maximum of 32.

\begin{table}[t!]
\centering
\small
\begin{tabular}{lr}
\hline
Dialogues             & 1,002 \\ 
Utterances / Emotion Transcripts            & 10,020 \\
Avg. Length per Utterance    & 42.72      \\
\hspace{2mm} by Speaker & 44.64 \\
\hspace{2mm} by Listener & 40.80 \\
Avg. Length per Emotion Transcripts  & 28.89 \\
\hspace{2mm} by Speaker  & 28.92 \\
\hspace{2mm} by Listener & 28.86 \\
\hline
\end{tabular}
\caption{Statistics of the Dataset}
\label{table:stat}
\end{table}

Table \ref{table:stat} shows key statistics of the collected dataset. On average, Speakers' utterances were slightly longer (in characters) than those of Listeners, while the lengths of emotional transcriptions were comparable across both roles.

\subsection{Annotation for Emotion Categorization}\label{sec:annotation_emotion_labels}
While the emotion transcriptions we collected contain fine-grained details of the speakers' emotional states, they are challenging to analyze quantitatively. To address this, we additionally annotated each emotion transcription in our dataset with corresponding emotion labels. This annotation not only enables a quantitative analysis of the emotions expressed in the transcriptions but also allows our dataset to be applied to traditional ERC tasks.\par

We adopted a set of seven emotion categories, comprising the six basic emotions from Ekman's six universal emotions ~\cite{ekman-etal-1987-universals}—joy, sadness, fear, anger, surprise, and disgust (喜び, 悲しみ, 恐怖, 怒り, 驚き, 嫌悪)—supplemented with a ``neutral''~(該当なし) category. This set is widely used in established ERC datasets~\citelr{Hsu2018EmotionLines,Poria2019meld,Li2017dailydialog}. To account for the possibility that a single emotion transcription may express multiple emotions, we employed a multi-label annotation approach. Specifically, annotators assigned all relevant labels to each emotion transcription. ``Neutral'' was used only for transcriptions that did not express any specific emotion.\par

Each emotion transcription was annotated by three different annotators on CrowdWorks. Annotators were provided with the full dialogue context and the corresponding emotion transcription, and made a binary answer on whether each of the seven emotion categories was present in the transcription. Following previous studies~\citelr{Busso2008iemocap, Hsu2018EmotionLines}, we used majority voting to aggregate the results; an emotion category was assigned as the final label if two or more of the three annotators agreed on it. If none of the emotion categories were agreed upon, the transcription was labeled as ``Neutral'' only.\par
\begin{table*}[t!]
\centering
\small
\renewcommand{\arraystretch}{1.3}
\scalebox{0.96}{
\begin{tabular}{lcccccccccccc}
\hline
\textbf{Models} & \textbf{Setting} & \textbf{B-1} & \textbf{B-2} & \textbf{B-3} & \textbf{B-4} & \textbf{R-1} & \textbf{R-2} & \textbf{R-L} & \textbf{BS} & \textbf{Prec.} & \textbf{Rec.} & \textbf{F1} \\
\hline
\multirow{2}{*}{GPT-4.1} 
& zero-shot & 16.89 & 7.99 & 3.93 & 2.09 & 23.61 & 4.87 & 17.93 & 57.66 & 14.73 & \textbf{42.27} & \underline{13.99}\\
& 4-shot & 26.89 & 13.34 & 7.36 & 4.12 & \underline{28.06} & \underline{5.78} & 22.98 & 59.67 & \underline{20.59} & \underline{27.40} & 13.78 \\
\hline 
\multirow{3}{*}{Llama-3.1}& zero-shot & 17.40 & 7.20 & 3.55 & 1.84 & 20.25 & 3.08 & 16.26 & 55.24 &  9.18 & 17.98 & 5.77\\
& 4-shot & \underline{29.41} & \underline{14.86} & \underline{8.46} & \underline{4.84} & 27.51 & 5.58 & \underline{23.22} & \underline{59.95} & 14.83 & 14.18 & 7.84\\
& fine-tuning & \textbf{36.07} & \textbf{23.01} & \textbf{15.59} & \textbf{9.98} & \textbf{31.95} & \textbf{8.79} & \textbf{28.54} & \textbf{62.64} & \textbf{28.50} & 19.71 & \textbf{14.29} \\
\hline
\end{tabular}
}
\caption{Automatic evaluation results for ETC task. Reported metrics include BLEU (B), ROUGE (R), BERTScore (BS), Precision (Prec.), Recall (Rec.), and F1-score (F1). Best values are in \textbf{bold}, and the second-best values in \underline{underlined}. }
\label{tab:autoeval_results}
\end{table*}

\begin{table}[t!]
\centering
\small
\renewcommand{\arraystretch}{1.3}
\scalebox{0.96}{
\begin{tabular}{lccccc}
\hline
\textbf{Model} & \textbf{Setting} & \textbf{\# Units (SD)} \\
\hline
\multirow{2}{*}{GPT-4.1} 
 & zero-shot & 3.39~(1.14) \\
 & 4-shot   & 1.96~(0.76) \\
 \hline
\multirow{3}{*}{Llama-3.1} 
 & zero-shot & 1.99~(1.04) \\
 & 4-shot & 1.58~(0.58) \\
 & fine-tuning & 1.39~(0.60) \\
\hline
\multirow{1}{*}{Reference} 
 & -- & 1.35~(0.64)\\
\hline
\end{tabular}
}
\caption{Average number of atomic units decomposed from the predicted transcription and from the ground truth~(Reference).}
\label{tab:autoeval_factscores}
\end{table}

\section{Experiments}
To evaluate the utility of our dataset for the ETC task, we trained baseline models and assessed their performance. We divided our dataset into training, validation, and test sets in an 8:1:1 ratio.

\subsection{Task}
The ETC task requires models to generate natural language descriptions reflecting a speaker's emotional state. To evaluate this ability, we formulate the ETC as follows. \par
Each dialogue $D$ in our dataset consists of $N=10$ utterances and is represented as a sequence of utterance-speaker pairs, denoted as $ D = ((u_1, s_1), (u_2, s_2), \dots, (u_N, s_N))$, where $u_i$ denotes the $i$-th utterance and $s_i$ represents its speaker.
In the ETC task, the model $\mathcal{M}$ is given the dialogue context $C_n$ up to the $n$-th utterance, denoted as $C_n = ((u_1, s_1), (u_2, s_2), \dots, (u_n, s_n))$,
where $n$ can be any utterance in the dialogue (i.e., $1 \le n \le N$). The objective of this task is to predict the emotion transcription $e_n$ for an utterance $u_n$ made by speaker $s_n$ based on $C_n$. i.e., $e_n = \mathcal{M} (C_n)$.
Here, the emotion transcription $e_n$ is a natural language description of the emotional state of speaker $s_n$ at the time of uttering $u_n$.\par

\subsection{Models}
We investigated the performance of two LLMs: \textbf{GPT-4.1}\footnote{We used gpt-4.1-2025-04-14.}, the latest advanced chat model developed by OpenAI, and \textbf{Llama-3.1-Swallow}\footnote{We used \href{https://huggingface.co/tokyotech-llm/Llama-3.1-Swallow-8B-Instruct-v0.3}{Llama-3.1-Swallow-8B-Instruct-v0.3}.} (\citealp{fujii2024continual}; \citealplr{okazaki2024building})
, an open-source LLM known for its strong proficiency in Japanese. Both LLMs were evaluated using zero-shot and 4-shot prompting.
The prompt template we designed for this experiment is presented in Appendix~\ref{sec:appx_prompt}.
Additionally, we performed supervised fine-tuning on Llama-3.1 with our dataset.
For the fine-tuned Llama-3.1, we report the average performance across models trained with five random seeds to ensure robustness. 
Further implementation details are provided in Appendix \ref{sec:appx_implementation}.

\subsection{Evaluation Metrics}\label{sec:eval_metrics}

\subsubsection{Traditional Automatic Evaluation Metrics}\label{sec:metric_traditional}
To evaluate the quality of the emotion transcriptions, we use two traditional automatic evaluation metrics: BLEU \cite{papineni2002bleu} and ROUGE \cite{lin-2004-rouge}, and a semantic similarity metric: BERTScore \cite{Zhang2020BERTScore}.\par

\subsubsection{Fine-grained and Interpretable Evaluation of Content Faithfulness}\label{sec:metric_factscore}
Recent studies have proposed using LLMs as automatic evaluators to approximate human preferences more closely \cite{zheng2023judging, liu-etal-2023-g}. Following this trend, we initially explored a LLM-based evaluation metric to assess the semantic alignment between the generated emotion transcriptions and the ground truth. 

However, relying on a single, holistic score presents significant challenges for our task. The emotional states are often complex, containing multiple components, such as ``\textit{feeling angry that my feelings are not understood, but also sad.}'' For such complex expressions, a single score struggles to accurately penalize the omission of some emotional components or the hallucination of others. To address this gap, we introduce a fine-grained evaluation approach of content faithfulness. This method, inspired by FActScore~\cite{min-etal-2023-factscore} is based on two steps.

\noindent\textbf{Decomposition into Atomic Units}. First, the source emotion transcription is decomposed into atomic units, each conveying a single piece of information about the emotional state. For example, \textit{feeling angry that my feelings are not understood, but also sad.}'' is split into two units: \textit{I am angry that my feelings are not understood}'' and ``\textit{I am sad that my feelings are not understood}.'' Note that some input texts may not contain any emotional description. We used Gemini-2.5-Flash\footnote{\url{https://ai.google.dev/gemini-api/docs/models}} for this step.\par

\noindent\textbf{Assessing Support for Atomic Units}. Next, we classify whether each atomic unit from the source transcription is supported by the target emotion transcription into three categories: \texttt{Supported}, \texttt{Not Supported}, and \texttt{Neutral}. For example, the unit ``I am angry that my feelings are not understood'' is supported by the target ``feeling angry that my feelings are not understood, but also sad.'' and is thus classified as \texttt{Supported}. In contrast, the unit ``I am happy that my feelings are understood'' is not supported by the target and is classified as \texttt{Not Supported}. The \texttt{Neutral} applies when a core component matches, but the validity of a secondary component is uncertain. For instance, relative to the target ``I am angry,'' the unit ``I am angry that my feelings are not understood'' is classified as \texttt{Neutral} because the core emotion (``angry'') matches, but the reason (``that my feelings are not understood'') cannot be verified. This classification step was performed using Gemini-2.5-Flash. 
The prompts used in the above two steps are presented in Appendix~\ref{sec:appx_llmjudge}.\par

Following these two steps, we compute two scores for each predicted emotion transcription.
\noindent\textbf{Precision}. We decomposed the predicted emotion transcription into atomic units in the first step and calculated the proportion of these units that are \texttt{Supported} by the ground truth transcription.\par
\noindent\textbf{Recall}. We decomposed the ground truth emotion transcription into atomic units and calculated the proportion of these units that are \texttt{Supported} by the predicted emotion transcription.\par
For both metrics, units classified as \texttt{Neutral} are not counted as correct in the calculation. We exclude cases where the ground-truth transcription does not describe any emotional state in our evaluation, as the above scores are undefined in such cases. If the generated transcription contains no emotional description, but the ground truth does, we set both scores to 0.
Additionally, we calculate the \textbf{F1-score}, the harmonic mean of Precision and Recall, as an overall measure of performance.

\begin{table*}[t!]
  \centering
  \small
  \scalebox{1.0}{
    \setlength{\tabcolsep}{2mm}
    \renewcommand{\arraystretch}{1.2}
    \begin{tabularx}{\linewidth}{lX}
      \toprule
      \textbf{Context}
      & \textbf{Speaker A}: The other day, I was driving at night when a cyclist with no lights suddenly shot out in front of me... It was terrifying.~(この間、夜に車を運転していたら無灯火の自転車が飛び出してきて…本当に怖い思いをしました。) \\
      & \textbf{Speaker B}: That sounds so scary. An experience like that could take years off your life.~(想像するだけで怖いです。寿命が縮みそうな出来事でしたね。)\\
      & \textbf{Speaker A}: I thought my heart was going to stop. I was relieved we didn't crash, but afterward, I just got so angry.~(心臓が止まりそうでしたよ、事故にならなくてほっとしましたが、あとから怒りが湧いてきましたね。)\\
      \midrule
      \textbf{Output}
      & \textbf{Ground-Truth}: I was happy that the dialogue partner understood my feelings, and I wanted to share my feelings even more.~(相手が心情を理解してくれたことが嬉しく、気持ちをもっときょうゆうしたくなりました。)
      \\[0.3em]
      & \textbf{GPT-4.1 (zero-shot)}: I'm feeling a mix of fear from the truly dangerous situation I was in, relief that it didn't turn into an accident, and anger that is now welling up toward the cyclist who shot out without lights. I'm speaking with the hope that the dialogue partner understands my shock and anger from that moment.~(自分が本当に危険な目にあったことへの恐怖と、事故にならずに済んだ安堵感、そして無灯火で飛び出してきた自転車に対して後から怒りがこみ上げてきた気持ちが混ざっている。相手にもそのときの自分の動揺や怒りを理解してほしいと思いながら話している。)
      \\[0.3em]
      & \textbf{GPT-4.1 (4-shot)}: I wanted the dialogue partner to understand my feelings of fear and anger.~(相手に自分の恐怖や怒りの気持ちを理解してもらいたかったです。)
      \\[0.3em]
      & \textbf{Llama-3.1 (zero-shot)}: I was really scared, and I also feel angry.~(本当に怖かったし、怒りも感じています。)
      \\[0.3em]
      & \textbf{Llama-3.1 (4-shot)}: I want the dialogue partner to understand my fear and anger.~(相手に自分の恐怖と怒りを理解してもらいたいと思っています。)
      \\[0.3em]
      & \textbf{Llama-3.1 (fine-tuned)}: I was happy for their empathy and wanted them to hear more details.~(相手の共感に嬉しく思い、さらに詳しく話を聞いてもらいたいと思いました。)
      \\
      \bottomrule
    \end{tabularx}
  }
  \caption{Case study of emotion transcription generation. (translated from Japanese) The outputs are the emotion transcriptions for Speaker A's final utterance.}
  \label{tab:case_study}
\end{table*}

\subsection{Results}

Table~\ref{tab:autoeval_results} presents the results of automatic evaluation for the ETC task. For widely used, traditional metrics — BLEU, ROUGE, and BERTScore — the fine-tuned Llama-3.1 model outperformed the other models by a significant margin.

In terms of Precision, the Llama-3.1 model fine-tuned on our dataset achieves the highest score of 28.50\%. For this metric, the 4-shot setting consistently outperforms the zero-shot setting, aligning with the trend observed in traditional automatic evaluation metrics. Recall, however, exhibits an opposite trend. The zero-shot setting yields higher scores than the 4-shot setting for both models, with the zero-shot GPT-4.1 achieving the highest Recall of 42.27\%—significantly outperforming all other models. Finally, for the F1-score, the fine-tuned Llama-3.1 model secures the best result at 14.29\%. Nevertheless, GPT-4.1 remains highly competitive, with its zero-shot setting achieving a close score of 13.99\%.

The fine-grained evaluation of content faithfulness reveals the distinct characteristics of each model. We present the average number of atomic units decomposed from the generated transcription and from the ground truth in Table~\ref{tab:autoeval_factscores}. The fine-tuned Llama-3.1 model generates transcriptions with an average atomic unit of 1.39, which is closest to the ground truth of 1.35 among all models. In contrast, models in the zero-shot setting, particularly the GPT variants, tend to generate higher numbers of atomic units per transcription. \par

The trends observed in the information content explain the Precision-Recall trade-off. Transcriptions with a higher number of atomic units, such as those from the zero-shot models, are more likely to cover the atomic units present in the ground truth, thereby enhancing their Recall scores. Conversely, this verbosity increases the likelihood of including redundant information, which negatively impacts their Precision. In contrast, the fine-tuned Llama-3.1 model's output quantity (1.39 units) not only aligns closely with the ground truth (1.35) but also achieves the highest Precision. This suggests that fine-tuning helped the model accurately identify the speaker's emotional state.\par

Moreover, the ETC task remains challenging, as evidenced by the overall low scores across all evaluation metrics. Even the best-performing model, the fine-tuned Llama-3.1, achieves only 14.29\% in F1-score, indicating significant room for improvement in accurately capturing speakers' emotional states in dialogues. Additionally, lower F1 scores than both Precision and Recall for all models suggest a notable imbalance between these two aspects, highlighting the difficulty of generating emotion transcriptions that are both comprehensive and precise. Bridging this gap positions our dataset as a valuable and challenging benchmark for the community, advancing future research on understanding emotional states.\par

\section{Case Study}
To illustrate the challenges faced by current ETC models, we present a case study in Table~\ref{tab:case_study}. In this example, Speaker A's final utterance explicitly mentions feelings of relief and anger related to a past event. The ground-truth transcription, however, reveals that the speaker's actual emotional state at that moment was happiness, stemming from the listener's empathetic response. This gap between the narrated emotions and the speaker's actual emotional state poses a significant challenge for the ETC task.\par

Faced with this challenge, most models failed to capture the speaker's true emotional state. As shown in Table~\ref{tab:case_study}, the outputs from GPT-4.1 and the zero-shot and 4-shot Llama-3.1 models focused only on the negative emotions explicitly mentioned in the narrative. For example, the generation from GPT-4.1 in the zero-shot setting, which is the most lengthy, not only identified the emotions present in the utterance but also elaborated on them to express a desire for the listener's understanding. However, it still failed to predict the emergent feeling of happiness. In contrast, the Llama-3.1 model, fine-tuned on our dataset, successfully inferred the speaker's happiness derived from the empathetic interaction with the listener, suggesting that our dataset provides valuable training signals for capturing nuanced emotional states in dialogues.\par

\newcolumntype{Y}{>{\raggedright\arraybackslash}X}

\section{Corpus Analysis}
The annotated emotion labels enable a structured analysis of our dataset, providing insights into the characteristics of the emotion transcriptions. In this section, we present a quantitative analysis of the annotated emotion labels, focusing on three aspects: the distribution and agreement of the labels~(Section~\ref{sec:label_dist_agree}), the co-occurrence of emotion labels within utterances~(Section~\ref{sec:cooccurrence}), and the textual differences between utterances and emotion transcriptions~(Section~\ref{sec:freq_words}).

\subsection{Label Distribution and Agreement}\label{sec:label_dist_agree}
We first analyze the distribution of emotion labels assigned to the emotion transcriptions, a process detailed in Section \ref{sec:annotation_emotion_labels}. Table~\ref{tab:emotion_freq_kappa} presents their distribution, showing the prevalence of each label for Speakers, Listeners, and all participants. In this analysis, prevalence is calculated as the proportion of transcriptions assigned a given label through majority voting. The table also reports the overall inter-annotator agreement, measured by Fleiss' kappa~\cite{fleiss-1971-measuring}.

Overall, approximately 50\% of the emotion transcriptions were labeled as ``Neutral,'' indicating that nearly half of the transcriptions expressed one or more emotions other than neutral. Among the emotion labels, ``Joy'' was the most frequently expressed.
When analyzed by role, Speakers' transcriptions have a higher proportion of emotion labels assigned compared to those of Listeners. This difference is likely to be attributed to the task instructions for each role. Speakers were explicitly instructed to talk about the topic related to a specific emotion, a prompt that naturally encouraged more emotional expression and recall, compared to the Listeners.
Interestingly, ``Surprise'' was the only emotion label expressed more frequently in the Listener's transcriptions than in the Speaker's (4.7\% vs. 2.4\%). Surprise is an emotion often experienced as a reaction to an event or new information. For example, our dataset contains several Listener transcriptions expressing surprise in response to the Speaker's narrative, such as ``\textit{It was somewhat unexpected, so I tried to imagine the situation myself.}'' and ``\textit{I was surprised that something like this could really happen.}'', thereby illustrating this trend.\par

The overall inter-annotator agreement of Fleiss' kappa is 0.533, indicating moderate agreement. Examining emotion labels individually, ``joy'' showed the highest agreement among annotators, followed by ``anger''~($\kappa=0.529$) and ``surprise.''~($\kappa=0.560$) Conversely, ``disgust'' exhibited the lowest agreement~($\kappa=0.233$).\par

\begin{table}[t!]
\centering
\small
\begin{tabular}{lrrrc}
\toprule
\textbf{Emotion} & \textbf{Spk. (\%)} & \textbf{Lsn. (\%)} & \textbf{All (\%)} & \textbf{$\kappa$} \\
\midrule
Joy & 29.9 & 20.0 & 25.0 & 0.603 \\
Sadness & 10.8 & 8.8 & 9.8 & 0.359 \\
Fear & 7.1 & 5.3 & 6.2 & 0.476\\
Anger & 3.7 & 2.8 & 3.2 & 0.529\\
Surprise & 2.4 & 4.7 & 3.5 & 0.560 \\
Disgust & 6.3 & 3.3 & 4.8 & 0.233 \\
Neutral & 43.4 & 57.2 & 50.3 & 0.400 \\
\midrule
Overall & -- & -- & -- & 0.533 \\
\bottomrule
\end{tabular}
\caption{Distribution of emotion labels in emotion transcriptions by dialogue role and inter-annotator agreement (Fleiss' kappa). Columns denote dialogue role: speaker~(Spk.), listener~(Lsn.), and both roles~(All).}
\label{tab:emotion_freq_kappa}
\end{table}

\begin{figure}[t!]
  \centering
  \includegraphics[width=\linewidth]{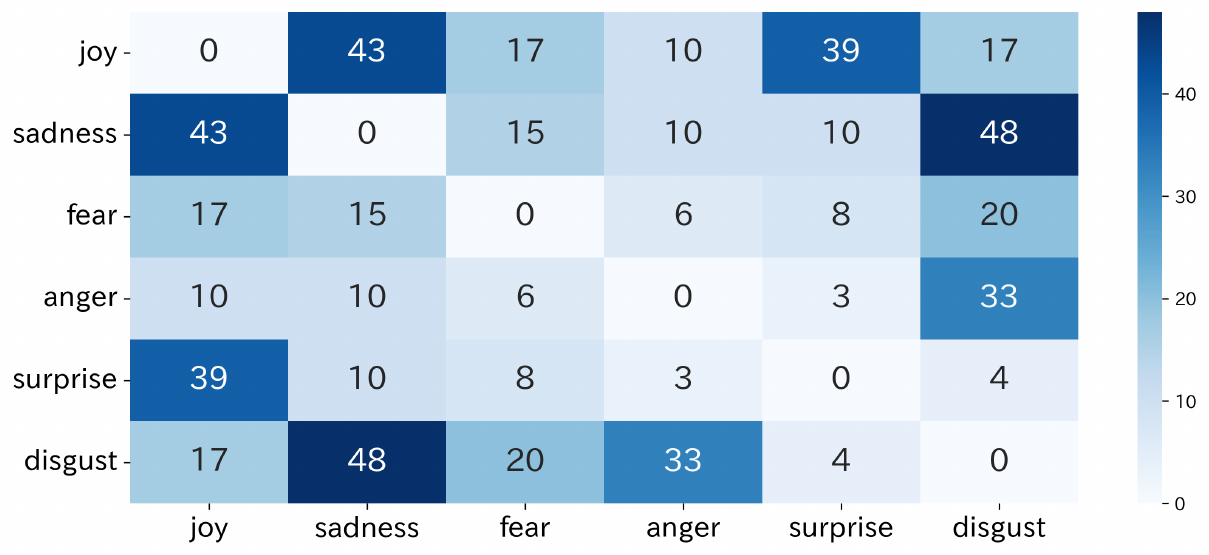}
  \caption{Co-occurrence matrix of emotion labels. Each cell indicates the number of transcriptions annotated with both corresponding labels.}
  \label{fig:emotion_cooccurrence}
\end{figure}

\newcolumntype{L}{>{\raggedright\arraybackslash}X}

\begin{table*}[t!]
\centering
\small
\begin{tabularx}{\textwidth}{@{}l L L @{}}
\toprule
\textbf{Emotion} & \textbf{Utterance} & \textbf{Emotion transcription} \\
\midrule

\textbf{Joy} &
くださる (give), 楽しみだ (excited), 感動 (moved), 食べる (eat), 楽しむ (enjoy) &
うれしい (happy), ワクワク (excited), 前向きだ (positive), 感動 (moved), 楽しみだ (excited) \\

\midrule
\textbf{Sadness} &
悲しい (sad), 寂しい (lonely), 辛い (painful), がっかり (disappointed), もう (enough) &
悲しい (sad), 残念だ (regretful), 寂しい (lonely), 辛い (painful), 同情 (sympathy) \\

\midrule
\textbf{Fear} &
ニュース (news), 恐ろしい (terrifying), 事件 (incident), 心配だ (worried), 地震 (earthquake) &
恐怖 (fear), 心配だ (worried), 恐ろしい (terrifying), 想像 (imagine), 心配 (worry) \\

\midrule
\textbf{Anger} &
イライラ (irritated), 怒る (angry), 立つ (stand), 悪い (bad), 怒り (anger) &
怒り (anger), イライラ (irritation), 強い (strong), いらいら (irritated), 怒る (angry) \\

\midrule
\textbf{Surprise} &
なんて (what), びっくり (surprised), くらい (about), 凄い (amazing), 時代 (times) &
驚き (surprise), 意外だ (unexpected), 感心 (impressed), びっくり (surprised), 想像 (imagine) \\

\midrule
\textbf{Disgust} &
嫌だ (dislike), 嫌悪 (disgust), 悪い (bad), 怒る (angry), 苦手だ (averse) &
嫌だ (dislike), 嫌悪 (disgust), 想像 (imagine), 抱く (harbor), 申し訳ない (sorry) \\

\midrule
\textbf{Neutral} &
あまり (little), くださる (give), 信頼 (trust), 食べる (eat), ところ (point) &
書く (write), 同意 (agree), 答える (answer), 具体 (specific), 説明 (explain) \\

\bottomrule
\end{tabularx}
\caption{Top-5 frequent words by category for utterances and emotion transcriptions.}
\label{tab:top-words-single}
\end{table*}

\subsection{Co-occurrence of Emotion Labels}\label{sec:cooccurrence}
In this section, we analyze the co-occurrence of emotion labels in the emotion transcriptions to understand the complexity of emotional expression. Figure \ref{fig:emotion_cooccurrence} illustrates the co-occurrence relationships of emotion labels within the transcriptions. We found that approximately 5.6\% of all non-neutral transcriptions contained multiple emotion labels.\par

A closer look at the co-occurrence matrix in Figure~\ref{fig:emotion_cooccurrence} reveals several patterns. First, emotions with similar valence frequently co-occur, such as ``Disgust'' with ``Anger'' and ``Sadness.'' Second, interestingly, emotions with opposing natures, like ``Joy'' and ``Sadness,'' also co-occur frequently. This co-occurrence is often not contradictory, as the two emotions can stem from different aspects of the situation. For example, transcriptions from the dataset included statements such as, ``I am recalling both pleasant and bitter memories.'' and ``listener is empathizing with me, but I also still feel sad when I recall the shopping scene.'' Furthermore, ``Disgust'' tends to co-occur broadly with a wide range of other emotions. This broad association may contribute to its ambiguity, potentially explaining why it received the lowest inter-annotator agreement. These results suggest the complexity of human emotion, which often involves multiple, sometimes conflicting feelings simultaneously.\par

\subsection{Frequent Words for Emotion Categories}\label{sec:freq_words}
In this section, to analyze the differences in linguistic features between utterances and emotion transcriptions, we extract and compare their respective characteristic words for each emotion label.
This analysis is based on the method conducted by \citetlr{ide-kawahara-2022-building}. Specifically, we first use Juman++~\cite{tolmachev-etal-2018-juman} to identify words appearing in an utterance or emotion transcription. Next, to filter out common words that appear across all emotion labels, we remove any word with an IDF value less than half of the maximum IDF. Finally, we extract the highest-frequency words as characteristic words.\par
Table \ref{tab:top-words-single} shows the Top-5 most frequent words for each emotion label. A common trend in both utterances and emotion transcriptions is the extraction of words that express the corresponding emotion category. For example, for joy, words like ``excited'' are frequent, and for sadness, ``sad'' is the most frequent.

A key difference emerged between the two text types. In utterances, the characteristic words tended to include not only emotion words but also words describing events and actions that trigger emotions. For instance, for joy, the word ``eat'' is extracted, while for fear, words like ``incident'' and ``earthquake'' are prominent. In contrast, the words that are frequent in emotion transcriptions tend to express the subtle nuances of the emotion. For example, in the category of joy, words like ``happy'' and ``excited'' appeared, while for fear, ``worried'' and ``terrifying'' are extracted. This shows a linguistic difference in expression even within the same emotion category.\par

For the neutral category has a trend of general words, rather than emotion words. Notably, for the emotion transcriptions, the words often used to describe an utterance's intent were extracted as characteristic words, ``agree,'' ``answer,'' and ``explain.''

\section{Conclusion and Future Work}
We introduced Emotion Transcription in Conversation (ETC), a novel task for generating verbalizations of a speaker's emotional states, capturing fine-grained emotional nuances. To support this task, we constructed a new dataset of 1k Japanese dialogues annotated with self-reported emotion transcriptions and benchmarked baseline models. Our experimental results indicate the utility of our dataset for the ETC task, while also highlighting the challenges. A key challenge is the potential gap between explicitly expressed emotions and the actual emotional state; even fine-tuned models failed to capture these nuances (see Appendix~\ref{sec:appx_failure} for failure cases). This suggests the need for models with enhanced capabilities to infer nuanced emotional states.\par

Future work can address this challenge from several directions. One is to explore sophisticated prompting (e.g., chain-of-thought) and advanced fine-tuning objectives, such as RLHF or contrastive learning, to distinguish subtle expressions. Furthermore, speaker modeling could enhance model capabilities. Building on prior ERC research, this approach could involve incorporating speaker embeddings~\cite{hu-etal-2021-mmgcn} or leveraging personality traits we have collected, which previous studies have linked to emotional states in dialogue~(\citealp{wang-etal-2024-emotion}; \citealplr{shinoda2025tomato}).\par

Additionally, designing human evaluation for the ETC task is a critical future direction. Assessing emotion transcriptions requires capturing subtle nuances in emotional expressions. While LLM-based evaluators have shown promise in this regard, their reliability and potential biases warrant careful examination. Comparing LLM-based and human evaluations would provide valuable insights into the validity of automatic evaluation for this task.

\section*{Limitations}

This study has the following limitations and challenges:
First, the ETC dataset was constructed using crowdsourcing with Japanese-speaking participants, which may introduce cultural and linguistic biases.
To confirm the generalizability of this dataset, similar data collection efforts across diverse languages and cultural contexts are required.
Second, our dataset is limited to the text modality, which may lack nonverbal cues essential for understanding humans' emotional states. Therefore, constructing multimodal datasets that incorporate audio and visual information is a promising direction for enriching emotion understanding in real-world conversations.
Third, the collected dialogues are relatively brief, comprising only 5 turns (10 utterances).
This short context might be insufficient to adequately capture longer conversational dynamics and the complexity of emotional states evolving over time.
Additionally, our dataset comprises 1k dialogues, and it remains unclear whether this scale is sufficient for training robust ETC models. However, the crowdsourcing-based collection protocol employed in this study provides a feasible and promising approach for scaling up ETC datasets in the future.
Lastly, additional consideration is necessary regarding potential use cases of emotional transcriptions (ET).
Future research should explore how recognized ET can be effectively integrated into conversational systems to inform interactions and enhance emotional intelligence.

\section*{Ethical Considerations}
This study involves the use of data that captures internal emotional states of individuals expressed through natural language, which necessitates careful ethical consideration regarding the sensitivity of emotional content and the potential risks associated with AI applications.
First, we collected the dialogue data and corresponding emotion transcriptions for each utterance through a Japanese crowdsourcing platform.
Prior to the data collection, the participants were provided with clear explanations of the study's purpose, the nature of the data being collected.
The participants were asked to consent to the data use before they engaged in the dialogue.
The collected data contains no personally identifiable information and was stored in an anonymized format.
Also, the proposed Emotion Transcription in Conversation (ETC) task aims to generate natural language descriptions of a speaker's internal emotional state based on conversational context. This technology holds potential for applications in dialogue systems, educational support, and counseling. However, if misused, it may raise serious ethical concerns, such as excessive inference of emotions, reinforcement of cultural biases, or surveillance and manipulation of emotional states.
To address these risks, we evaluated the alignment between model-generated outputs and human annotations, and highlighted the current limitations and areas for improvement.
Nevertheless, continued attention must be paid to the ethical implications of how such trained models are deployed and used in real-world contexts.

\section*{Acknowledgments}
This work was supported by JSPS KAKENHI Grant Number 25H01382.

\section{Bibliographical References}\label{sec:reference}
\bibliographystyle{lrec2026-natbib}
\bibliography{custom}

\section{Language Resource References}\label{lr:ref}
\bibliographystylelr{lrec2026-natbib}
\bibliographylr{resource}

\clearpage
\appendix

\begin{figure*}[t!]
    \centering
    \includegraphics[width=1.0\linewidth]{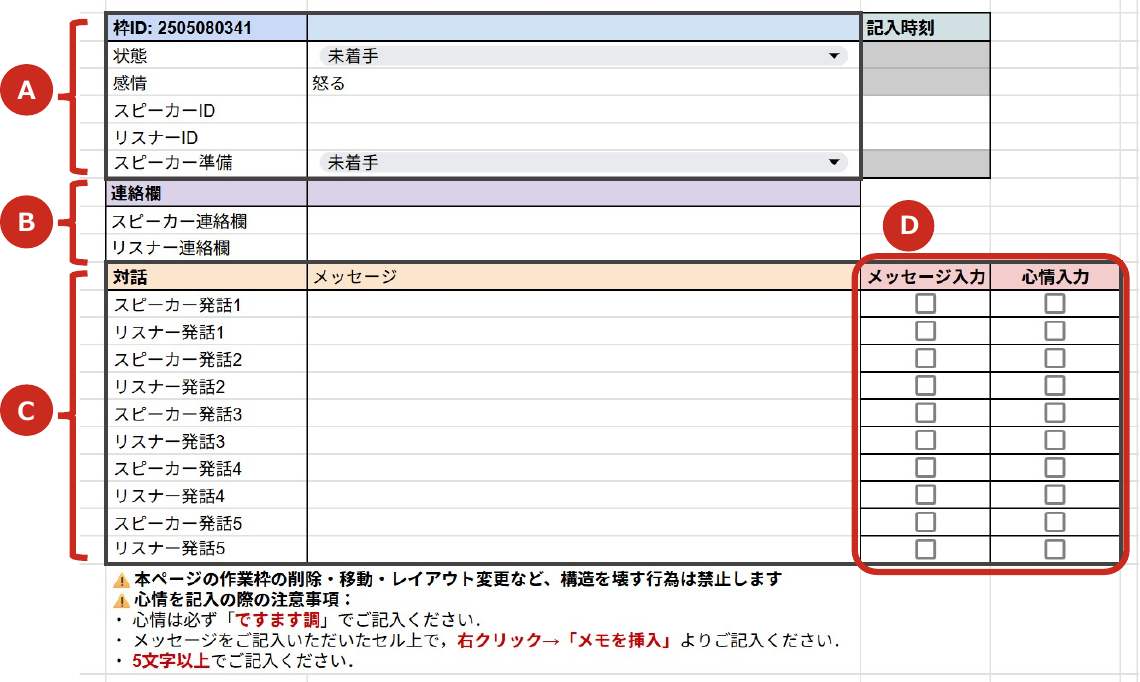}
    \caption{The dialogue slot used for conducting conversations and entering emotion transcriptions.}
    \label{fig:interface}
\end{figure*}

\section{Data Collection Details}
\label{sec:appx_collection}
We conducted the collection of dialogue and corresponding emotion transcriptions using a custom-built chat interface hosted on Google Sheets. The interface consists of multiple dialogue slots (as shown in Figure \ref{fig:interface}), each associated with a predefined emotion label.\par
Each dialogue slot in the interface is composed of several key areas. Figure~\ref{fig:interface}-A shows metadata for the dialogue, including a predefined emotion label and a unique dialogue slot ID. This area also includes two cells for entering CrowdWorks usernames—one for the Speaker and one for the Listener. In these cells, participants were instructed to enter their username to join the conversation. They then wait to be paired with a dialogue partner or start the conversation if someone has already joined.
Figure~\ref{fig:interface}-B allows participants to exchange coordination messages during the task. This area is designed for task-related communication that does not directly affect the dialogue itself and facilitates the accurate and smooth progression of tasks. For instance, participants used it to confirm matching by entering ``\textit{Nice to meet you}'', inform dialogue partners of their readiness with ``\textit{I'm ready}'', or notify other workers of a brief absence while waiting for matching with ``\textit{I'll be back by 4:00 PM}''.

Figure~\ref{fig:interface}-C is the main body of the interaction, where participants input their utterances across five turns each. After a participant (e.g., the Speaker) enters their utterance into a cell in this area, the other participant (e.g., the Listener) responds by entering their utterance in the next cell. Subsequently, while the Listener is typing their response, the Speaker writes an emotion transcription for their previous utterance by inserting a note into the same cell. These steps are repeated alternately for five turns. During the dialogue session, we required participants to write at least five characters in length for both the utterance and the corresponding emotion transcription.\par
Figure~\ref{fig:interface}-D consists of checkboxes for tracking task progress. The left column is for utterance input, and the right column is for emotion transcription. At each turn, participants check the corresponding box after completing each action. The dialogue session is considered complete only when all checkboxes have been marked by both participants.

\begin{figure*}[t!]
    \centering
    \includegraphics[width=0.8\linewidth]{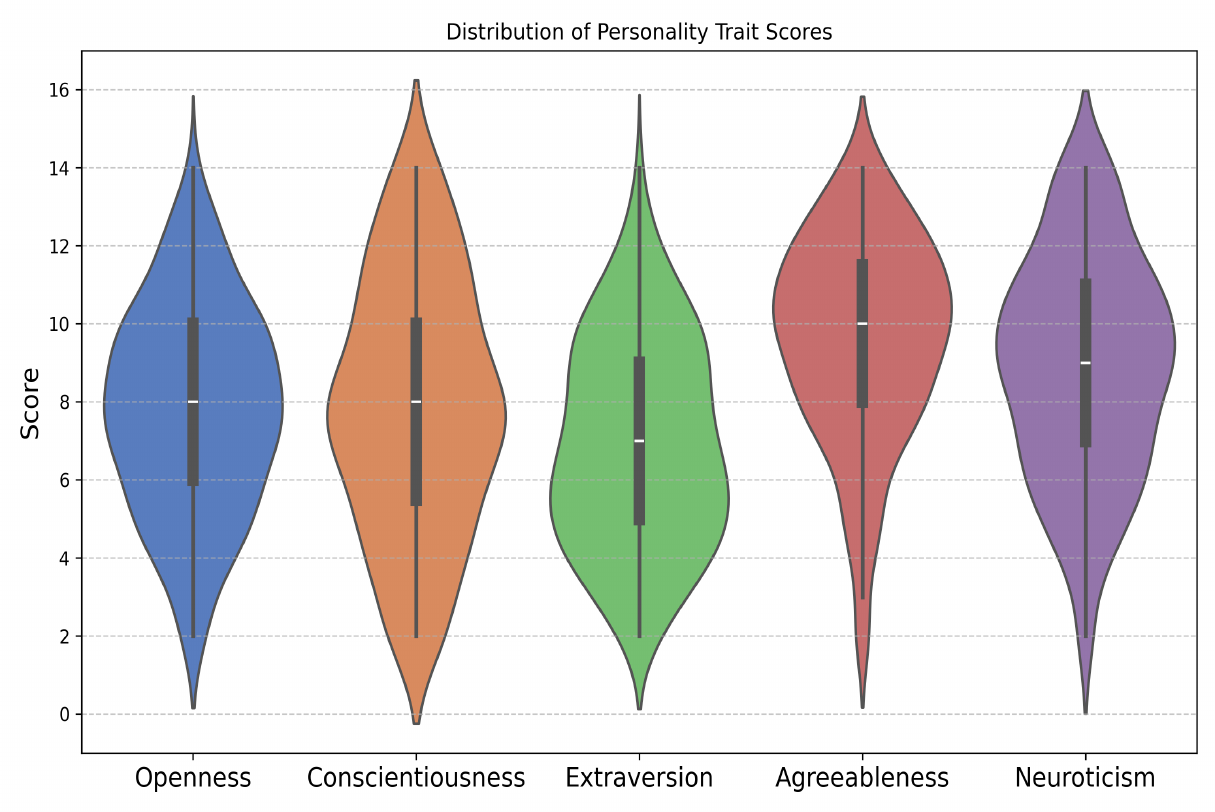}
    \caption{Distribution of Big Five personality traits among the 199 participants.}
    \label{fig:big5}
\end{figure*}

\begin{figure*}[t!]
    \centering
    \setlength{\fboxsep}{0pt} 
    \setlength{\fboxrule}{0.2pt} 
    \fbox{%
        \includegraphics[width=1.0\linewidth]{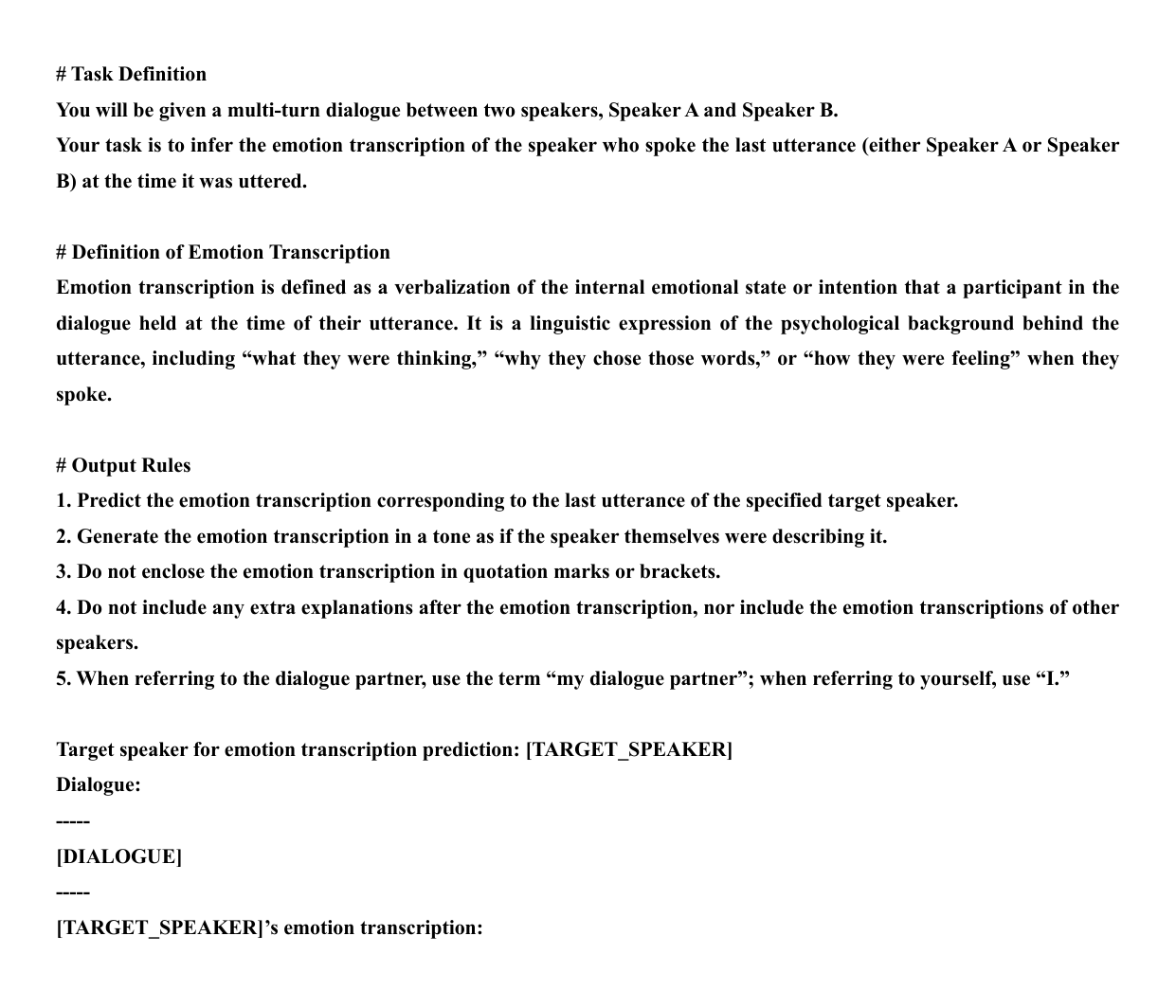}
    }
    \caption{The prompt template for the ETC task (Translated from Japanese). The model is expected to generate only the emotion transcription.}
    \label{fig:prompt_etc}
\end{figure*}

\begin{figure*}[t!]
    \centering
    \setlength{\fboxsep}{0pt} 
    \setlength{\fboxrule}{0.2pt} 
    \fbox{%
        \includegraphics[width=1.0\linewidth]{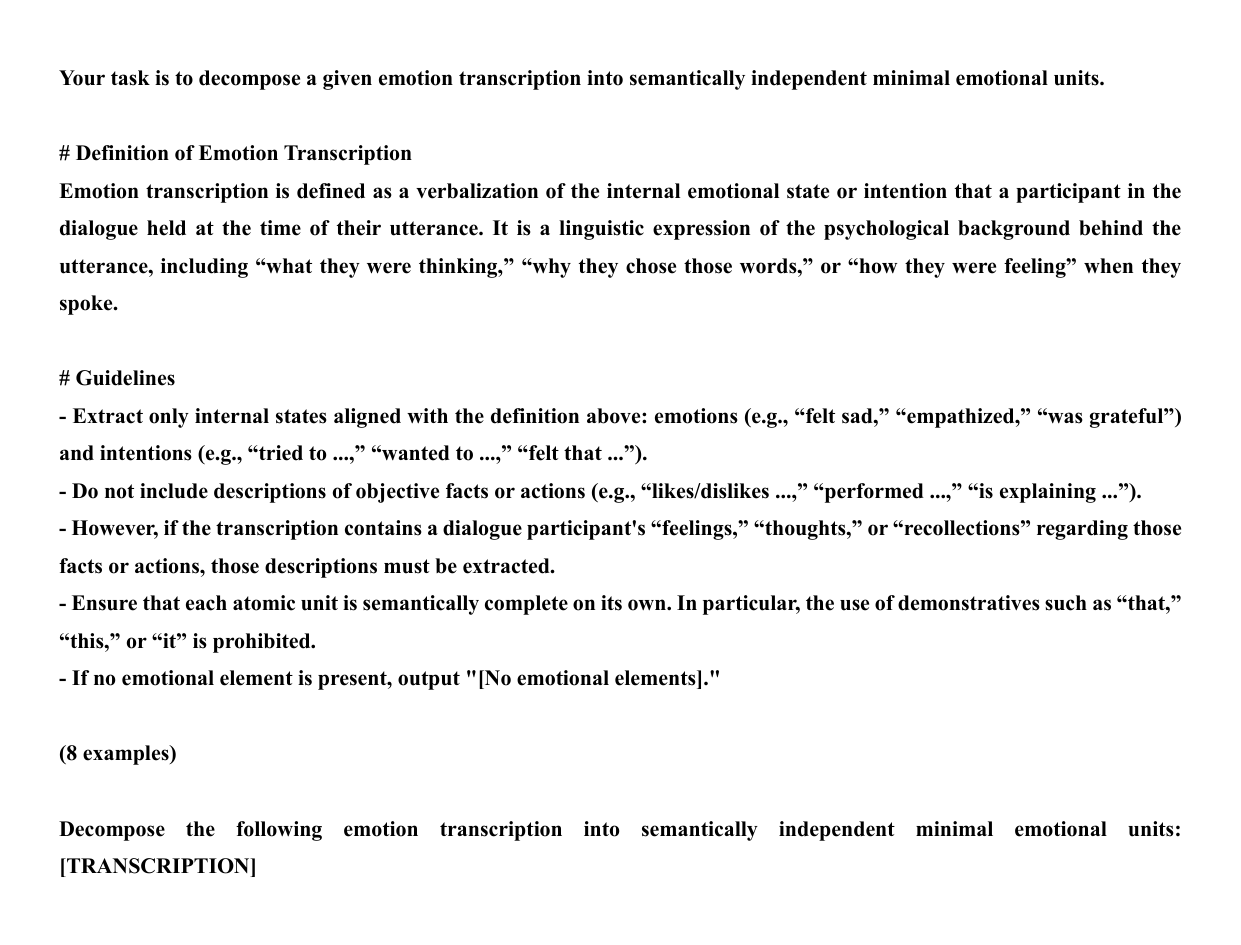}
    }
    \caption{The prompt template for the decomposition step in the fine-grained evaluation (Translated from Japanese)}
    \label{fig:prompt_decomp}
\end{figure*}

\begin{figure*}[t!]
    \centering
    \setlength{\fboxsep}{0pt} 
    \setlength{\fboxrule}{0.2pt} 
    \fbox{%
        \includegraphics[width=1.0\linewidth]{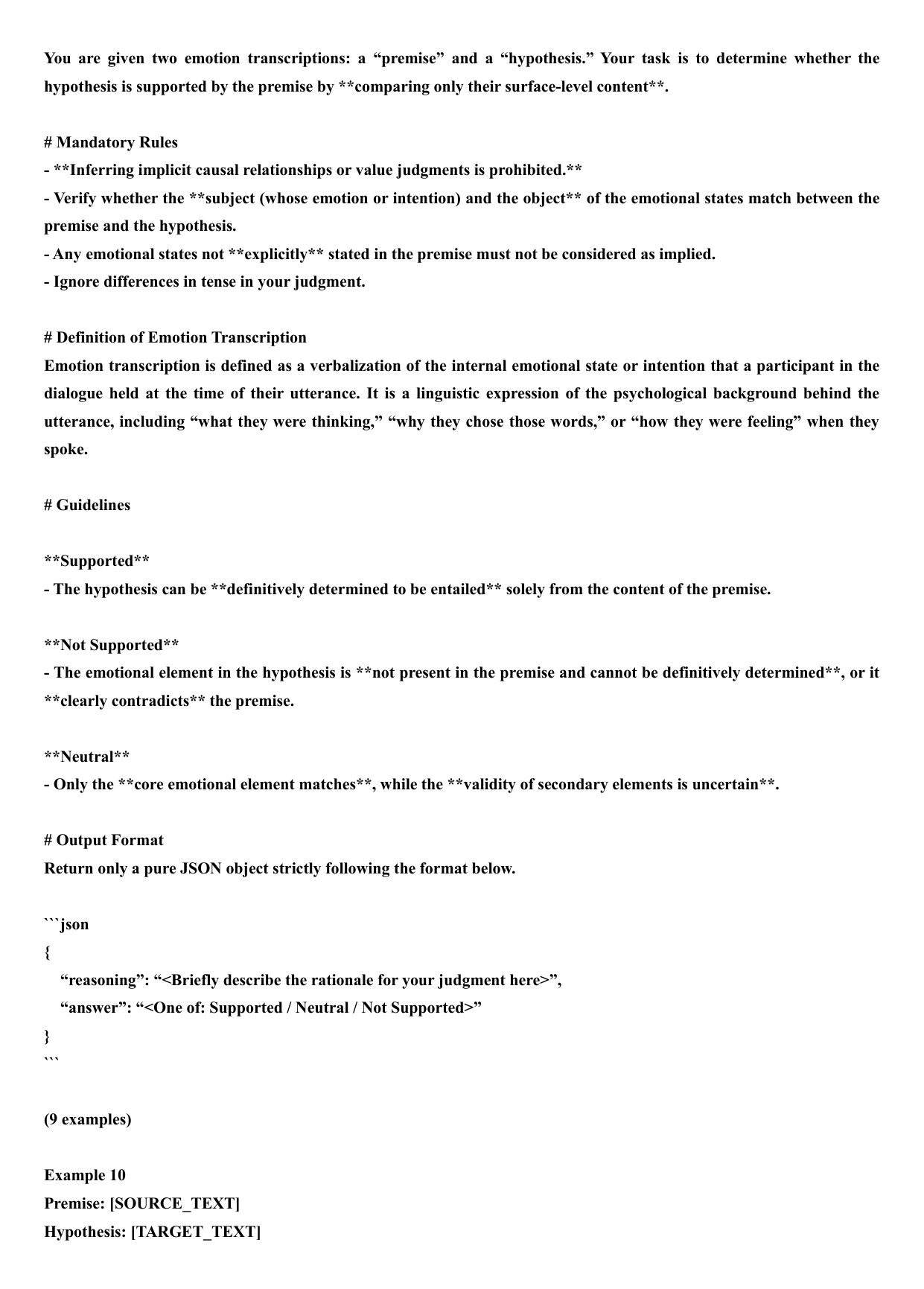}
    }
    \caption{The prompt template for the entailment step in the fine-grained evaluation (Translated from Japanese).}
    \label{fig:prompt_entail}
\end{figure*}

\section{Distribution of Participants' Personality Traits}
\label{sec:appx_personality}
Figure~\ref{fig:big5} illustrates the distribution of Big Five personality traits among the 199 participants, measured using the Japanese version of the Ten-Item Personality Inventory (TIPI-J)~\cite{oshio2012development}. Scores for each dimension range from 2 to 14, with higher values indicating stronger tendencies.

\section{Prompt Template for ETC Task}
\label{sec:appx_prompt}
The prompt template for the ETC task, translated from Japanese, is shown in Figure \ref{fig:prompt_etc}. The model is expected to generate only the emotion transcription, in accordance with the definition. \texttt{[TARGET\_SPEAKER]} is a placeholder for the target speaker whose emotion transcription is to be generated, and \texttt{[DIALOGUE]} is a placeholder for the dialogue history.

\section{Implementation Details}
\label{sec:appx_implementation}
We executed supervised fine-tuning of Llama-3.1-Swallow on two NVIDIA A100 80GB GPUs, using a batch size of \texttt{8} and training for \texttt{2} epochs. We applied 4-bit quantization to enhance memory efficiency and used the \texttt{PagedAdamW8bit} optimizer during training. Validation was conducted every \texttt{200} steps.
For hyperparameter tuning, we employed a grid search with learning rates tested in \texttt{\{1e-05, 5e-05\}} and warm-up steps in \texttt{\{300, 700\}}. As a result, we selected a learning rate of \texttt{1e-05} and \texttt{300} warm-up steps. \par
During the inference, to ensure reproducibility, we set \texttt{do\_sample=False} for Llama-3.1-Swallow and \texttt{temperature} to \texttt{0.0} for GPT-4.1.

\section{Prompt Template for Fine-grained Evaluation}
\label{sec:appx_llmjudge}
The prompt templates for the fine-grained evaluation of content faithfulness, translated from Japanese, are shown in Figure \ref{fig:prompt_decomp} and Figure \ref{fig:prompt_entail}. The decomposition prompt (Figure \ref{fig:prompt_decomp}) instructs the model to output a list of atomic units of information contained in the emotion transcription. \texttt{[TRANSCRIPTION]} is a placeholder for the emotion transcription to be decomposed. The entailment prompt (Figure \ref{fig:prompt_entail}) asks the model to determine whether each unit of information is entailed by the target transcription. \texttt{[SOURCE\_TEXT]} and \texttt{[TARGET\_TEXT]} are placeholders for the source and target emotion transcriptions, respectively.

\section{Failure Case Analysis for the ETC Task}
\label{sec:appx_failure}
Table \ref{tab:failure_case_study} presents a cherry-picked example of where the models failed to predict the speaker's emotion transcription. In this case, the ground-truth transcription expresses the speaker's delight at the partner's willingness to delve into details. However, every model instead focuses on the speaker's shock, illustrating a tendency to prioritize immediately preceding negative cues over more nuanced, interaction-driven emotions. This suggests that ETC models struggle to incorporate the partner's supportive attitude when inferring a speaker's emotional state.

\begin{table*}[t!]
  \centering
  \small
  \scalebox{1.0}{
    \setlength{\tabcolsep}{2mm}
    \renewcommand{\arraystretch}{1.2}
    \begin{tabularx}{\linewidth}{lX}
      \toprule
      \textbf{Context}
      & \textbf{Speaker A}: The other day, my favorite baseball player suffered a serious injury and will be out for three months.~(先日、自分が特に応援している野球選手が全治3ヵ月の大怪我を負ってしまいました。) \\
      & \textbf{Speaker B}: Three months? Are you talking about Okamoto?~(全治3ヶ月というと、それはもしかして岡本選手ですか？)\\
      & \textbf{Speaker A}: Yes. That news was really shocking.~(そうです。あの出来事は本当にショックでした。)\\
      & \textbf{Speaker B}: I thought so. He got hurt in that collision with Nakano from the Hanshin Tigers, and I know how devastating that is for a Giants fan.~(やっぱりそうですか。阪神の中野選手との交錯でケガをしたと思いますが巨人ファンの方からすると彼がいなくなるのは大ダメージですね。)\\
      & \textbf{Speaker A}: Right. At first they reported it as a one-month muscle injury, but after a detailed examination it turned out to be a ligament tear requiring three months of recovery, which made it even more shocking.~(はい。最初は筋肉の損傷で全治1カ月と報道されていたのにその後、精密検査で靭帯損傷で全治3ヵ月となったので、余計にショックでしたね。)\\
      \midrule
      \textbf{Output}
      & \textbf{Ground-Truth}: I feel very happy that my dialogue partner could dive so deeply and discuss the details with me.~(相手がかなり話を掘り下げた詳しいことまで話せることがとても嬉しいと感じています。)
      \\[0.3em]
      & \textbf{GPT-4.1 (zero-shot)}: I feel even more disappointed and shocked because the injury was worse than initially reported, and I'm explaining the details so they understand how bad it really is.~(自分が応援している選手のケガが思ったよりも重かったことに対して、最初の報道よりも悪い結果になってしまい、残念な気持ちやショックがさらに大きくなったと感じている。相手にその気持ちを理解してもらいたくて、詳しく説明している。)
      \\[0.3em]
      & \textbf{GPT-4.1 (4-shot)}: I wanted my partner to understand how shocked I was.~(相手に自分のショックな気持ちを理解してもらいたかったです。)
      \\[0.3em]
      & \textbf{Llama-3.1 (zero-shot)}: That incident was really shocking.~(あの出来事は本当にショックでした。)
      \\[0.3em]
      & \textbf{Llama-3.1 (4-shot)}: I feel relieved that my partner understands my feelings and I'm seeking empathy.~(相手が自分の気持ちを理解してくれていることに安心し、共感を求めています。)
      \\[0.3em]
      & \textbf{Llama-3.1 (fine-tuned)}: I answered their question, recalled what happened, and told them how much more shocked I felt.~(相手の質問に答えつつ、当時の状況を思い出して、さらにショックを受けたことを伝えました。)
      \\
      \bottomrule
    \end{tabularx}
  }
  \caption{Failure case study of emotion transcription generation (translated from Japanese). The outputs are the emotion transcriptions for Speaker A's final utterance.}
  \label{tab:failure_case_study}
\end{table*}








\end{document}